\documentclass[conference]{IEEEtran}
\IEEEoverridecommandlockouts
\usepackage{cite}
\usepackage{amsmath,amssymb,amsfonts}
\usepackage{algorithmic}
\usepackage{graphicx}
\usepackage{url}
\usepackage{textcomp}
\usepackage{xcolor}
\usepackage{cleveref}  
\usepackage{subcaption}
\usepackage{caption}
\def\BibTeX{{\rm B\kern-.05em{\sc i\kern-.025em b}\kern-.08em
    T\kern-.1667em\lower.7ex\hbox{E}\kern-.125emX}}
\begin{document}

\title{Copilot-Assisted Second-Thought Framework for Brain-to-Robot Hand Motion Decoding
\thanks{
\textsuperscript{†}These authors contributed equally to this work}
}

\author{\IEEEauthorblockN{
Yizhe Li\textsuperscript{†}
}
\IEEEauthorblockA{
\textit{University of Birmingham}\\
yxl1897@alumni.bham.ac.uk}
\and
\IEEEauthorblockN{
Shixiao Wang\textsuperscript{†}
}
\IEEEauthorblockA{
\textit{University of Birmingham}\\
sxw1238@student.bham.ac.uk}
\and
\IEEEauthorblockN{
Jian K. Liu
}
\IEEEauthorblockA{
\textit{University of Birmingham}\\
j.liu.22@bham.ac.uk}

}

\maketitle

\begin{abstract}
Motor kinematics prediction (MKP) from electroencephalography (EEG) is a key research area for developing movement-related brain–computer interfaces (BCIs). While traditional methods often rely on CNNs or RNNs, Transformers have demonstrated superior capabilities for modeling long sequential EEG data. In this study, we propose a CNN–attention hybrid model for decoding hand kinematics from EEG during grasp-and-lift tasks, achieving superb performance in within-subject experiments. We further extend this approach to EEG–EMG multimodal decoding, yielding significantly improved performance. Within-subject tests achieve PCC values of 0.9854, 0.9946, and 0.9065 (X, Y, Z), respectively, computed on the midpoint trajectory between the thumb and index finger, while cross-subject tests result in 0.9643, 0.9795, and 0.5852. The decoded trajectories from both modalities were used to control a Franka Panda robotic arm in a MuJoCo simulation. To enhance trajectory fidelity, we introduce a copilot framework that filters low-confidence decoded points using a motion-state-aware critic within a finite-state machine. This post-processing step improves the overall within-subject PCC of EEG-only decoding to 0.93 while excluding fewer than 20\% of the data points. The code is available at https://github.com/ssshiFang/EEGDecoding$\_$Robotarm.

\end{abstract}


\section{Introduction}
Electroencephalography (EEG) is a fundamental tool in neuroscience, offering the potential for controlling external devices directly through brain activity, bypassing the neuromuscular pathway \cite{wolpaw2012bci}. This concept is realized through brain–computer interfaces (BCIs). The central nervous system (CNS) typically generates movement in response to external stimuli, acting on effectors such as muscles or glands \cite{BrodalPer2010Tcns, yin_distributed_2025}. BCIs introduce an alternative pathway for this interaction by leveraging digital technologies. They can utilize signals from task-related brain regions to enhance, monitor, or even partially replace traditional CNS–effector communication \cite{PMID:32164849}.

While BCIs can theoretically use any brain-derived signal as input, practical limitations exist, such as random mental activity or external noise. Designing an effective BCI paradigm is therefore critical. By using various types of stimuli and tasks, a mapping can be established between specific inputs and corresponding brain activity patterns \cite{yu_toward_2020}. Current BCI systems are primarily built on paradigms like motor imagery, event-related potentials, and steady-state visually evoked potentials, all of which heavily rely on non-invasive EEG for data acquisition \cite{LeeMin-Ho2019EdaO}.

Decoding algorithms are another crucial component, translating measured brain signals into executable control commands. However, this step is particularly challenging due to the inherent properties of EEG. Scalp EEG represents a summation of currents from multiple local field potentials, reflecting broad brain activity \cite{10.5555/2424025, article}. Furthermore, interference between current sources complicates spatial localization, and signal transmission through various tissue layers results in a very low signal-to-noise ratio.
The advent of deep learning models, such as Shallow ConvNet and EEGNet, has demonstrated superiority over traditional handcrafted feature methods \cite{deOliveiraIagoHenrique2023Ecod, SchirrmeisterRobinTibor2017Dlwc, LawhernVernonJ2018Eacc}. Transformers have further shown powerful capabilities in handling long sequences, overcoming limitations of CNNs and RNNs \cite{IngolfssonThorirMar2020EAAT, song2021transformerbasedspatialtemporalfeaturelearning, CisottoGiulia2020CoAD}. Consequently, hybrid models combining these architectures are increasingly applied to classification tasks \cite{EEGconformer, WangDanjie2025SAMC}. Despite these advances, most kinematic trajectory prediction tasks using movement-related EEG, beyond the core paradigm classifications, still rely on filter-based, linear, or CNN/RNN approaches \cite{KoblerReinmarJ2020Dask} and continue to struggle with decoding accuracy \cite{TRAMMEL2023120268}.

This work proposes a CNN–attention hybrid model for EEG-based hand kinematics reconstruction. We further investigate EEG–EMG multimodal decoding, reconstruct motion trajectories in a MuJoCo robotic arm simulation, and design a copilot algorithm to refine decoding points through secondary evaluation.

\section{Related Work}
EEG-based kinematics decoding aims to reconstruct continuous limb movement trajectories (e.g., position, velocity) from non-invasive EEG signals. Prior studies have targeted various body parts, including ankle plantar flexion \cite{JochumsenMads2013Daco}, hand kinematics \cite{Jain_2023}, finger movements \cite{PaekAndrewY.2014Drfm}, and even saccadic eye movements \cite{JiaYingxin2019Mose}. These tasks are closely linked to the temporal–spectral dynamics of EEG and have been explored in both motor imagery and motor execution paradigms \cite{ChaddadAhmad2023ESPA, EratKübra2024ErwE}. Among these, hand-movement decoding is the most extensively studied.

Early work often assumed EEG lacked sufficient information for complex hand movement representation, a notion later challenged by \cite{BradberryTrentJ2010RTHM}. \cite{OfnerPatrick2017Ulmc} further elucidated the role of contralateral motor cortical regions in upper-limb movement control. These studies laid the foundation for hand-kinematics decoding and advanced understanding beyond simple single-region activation models.


Deep learning models have demonstrated superior performance in capturing nonlinear relationships between EEG features and motor parameters, especially when combined with time–frequency or spatial–spectral preprocessing. A CNN–LSTM model decoded hand kinematics from cortical sources with a mean correlation value (CV) of 0.62\textpm0.08 \cite{Jain_2023}. For other body parts, biceps curl trajectory estimation reached a PCC of 0.7 \cite{SainiManali2024BPEN}, and a multi-directional CNN-BiLSTM network for 3D arm tasks achieved a grand-averaged correlation coefficient (CC) of 0.47 \cite{JeongJi-Hoon2020BRAS}. A comparison in \cite{JainAnant2024EESI} showed deep learning models (rEEGNet, rDeepConvNet, rShallowConvNet) significantly outperformed mLR, with mean PCCs reaching 0.78–0.79 for x/y axes and 0.62–0.64 for the z-axis.

Despite progress, emerging technologies offer new solutions. Hybrid CNN–transformer architectures have shown strong capabilities in EEG decoding tasks, such as EEGformer \cite{WanZhijiang2023EAtb}, EASM \cite{SinghMadan2025EAeA}, EEG-TCNTransformer \cite{signals5030034}, and EEG-ConvTransformerNetwork \cite{bagchi2021eegconvtransformersingletrialeegbased}. Given that most kinematics decoding models remain CNN- or RNN-based, the success of the AI copilot framework in \cite{0Brain} motivated our exploration of hybrid modeling and trajectory optimization for non-invasive EEG-based hand decoding.

The contributions of this study are as follows:
(1) We propose a CNN–attention hybrid model to predict 3D coordinates of the index finger and thumb from EEG. The model achieves an overall PCC of 0.8728 in within-subject experiments (0.8376, 0.9229, 0.7208 for X, Y, Z axes), reaching up to 0.8916 with a larger input window.
(2) We investigate EEG–EMG multimodal fusion for kinematics decoding. The best within-subject PCC is 0.9707 (0.9854, 0.9946, 0.9065 for X, Y, Z), and the best cross-subject total PCC reaches 0.9063 (0.9643, 0.9795, 0.5852 for X, Y, Z).
(3) We reconstruct hand movement trajectories in a MuJoCo robotic arm simulator and design a copilot framework based on a state machine and knowledge graph to filter unreliable decoding points.

\section{Materials and Methods}
\subsection{Dataset Description \cite{luciw_jarocka_edin_2014}}
The WAY-EEG-GAL (Wearable interfaces for hand function recovery EEG grasp and lift) dataset contains scalp EEG recordings from a grasp-and-lift task designed to decode sensory, intentional, and motor-related signals. Twelve participants each performed 328 trials, totaling 3,936 trials.

Each trial required participants to: (1) reach for a small object upon cue, (2) grasp it with the thumb and index finger, (3) lift and hold it briefly, (4) place it back, and (5) return the hand to the start position. The dataset includes 32-channel EEG, electromyography (EMG) from five muscles, and contact force/torque and 3D position data for the hand (index finger, thumb, wrist) and the object.

\begin{figure}[t]
  \centering
  \includegraphics[width=1\linewidth]{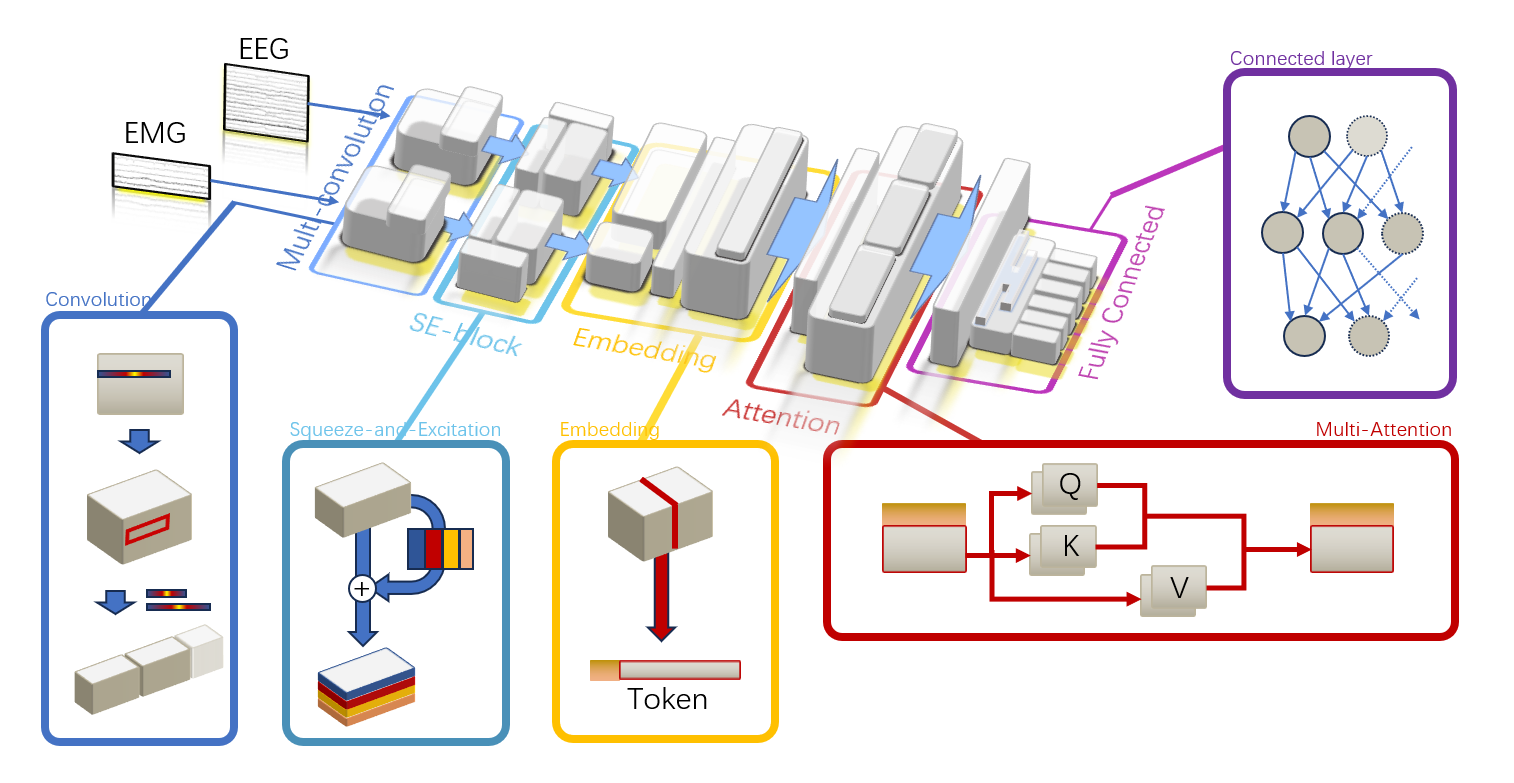}
  \caption{Structure of the decoding model.} 
  \label{fig:model}
\end{figure}

\subsection{Data Preprocessing}
EEG preprocessing was performed using MNE-Python. Raw data were band-pass filtered between 0.1 and 40 Hz with an infinite impulse response (IIR) filter to remove low-frequency drifts and high-frequency noise. Common average referencing (CAR) was then applied across all electrodes to reduce spatially distributed artifacts and enhance localized neural activity.

EMG preprocessing involved two steps. First, a 4th-order Butterworth band-pass filter (20–450 Hz), implemented in second-order sections (SOS) for stability, removed high-frequency noise and baseline drift. Second, EMG data were downsampled from 4000 Hz to 500 Hz to match the EEG sampling rate.

Finally, kinematic data were scaled via min–max normalization, mapping the processed data to the range [0, 1] as
$
K[t] = \frac{k[t] - k_{\text{min}}}{k_{\text{max}} - k_{\text{min}}}.
$

\subsection{Algorithm Structure}
\subsubsection{Model Overview}
The model architecture is illustrated in Fig.~\ref{fig:model}. It consists of five main components: a multi-convolution block, a Squeeze-and-Excitation (SE) block, an embedding block, a self-attention block, and a fully connected block. The input is a preprocessed EEG segment in the form of a 2D matrix (channels $\times$ time points).

\subsubsection{Multi-Convolution Block}
\begin{figure}[t]
  \centering
  \begin{minipage}{1\linewidth}
    \centering
    \includegraphics[width=1\linewidth]{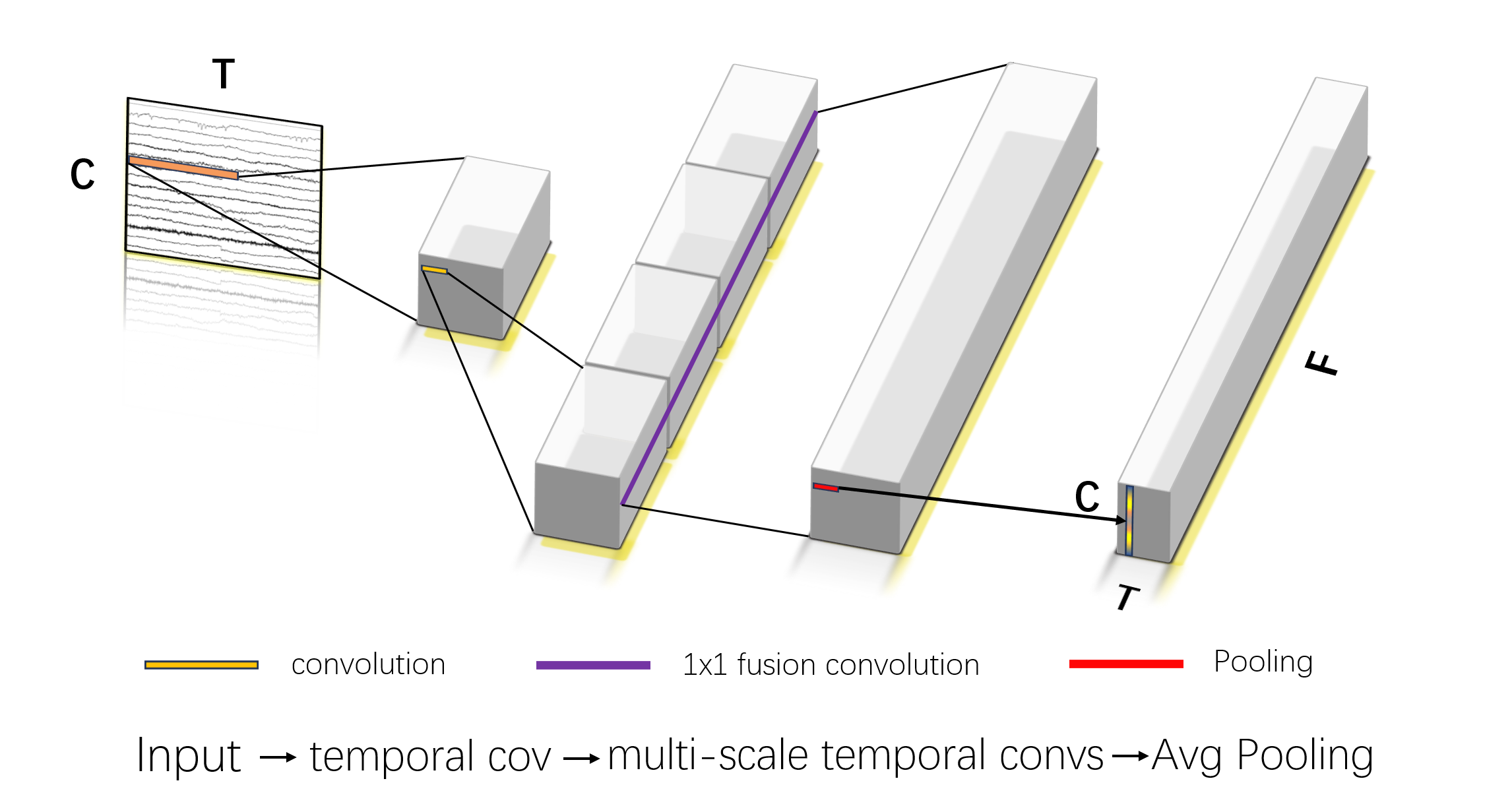}\\[0.5em]
    \includegraphics[width=1\linewidth]{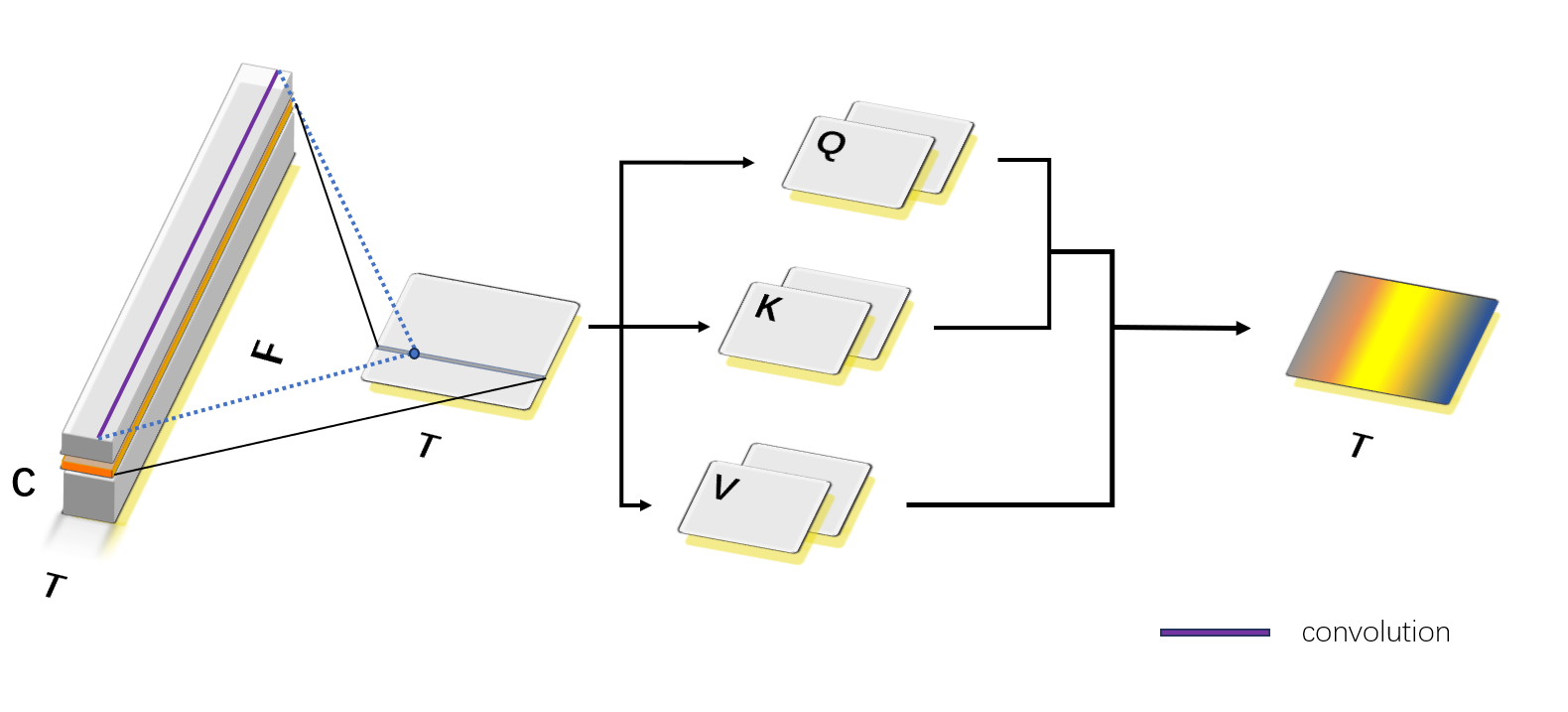}
    \captionof{figure}{Structure of the multi-convolution block (top). Structure of the self-attention block (bottom).}
    \label{fig:attention}
  \end{minipage}
\end{figure}

The multi-convolution block (Fig.~\ref{fig:attention}) is partially inspired by EEGNet. First, a 1D convolution with a large temporal kernel captures long-range dependencies \cite{BaiShaojie2018AEEo}. Subsequent layers employ multiple smaller kernels of varying sizes to extract multi-scale temporal features. The 1D convolution is implemented as
$
y[t] = \sum_{k=0}^{K-1} w[k] \cdot x[t + k],
$
where $y[t]$ is the output for the $t$-th element. A pooling layer then simplifies features and reduces computational cost for the subsequent attention mechanism. The average pooling operation is defined as
$
y_i = \frac{1}{k} \sum_{j=0}^{k-1} x_{i \cdot s + j}.
$

\subsubsection{Squeeze-and-Excitation (SE) Block}
The SE block highlights informative input channels. Following \cite{HuJie2020SN}, for an input feature map $\mathbf{X} \in \mathbb{R}^{B \times C \times F \times T}$, the block performs:
\begin{align}
\textbf{Squeeze:}    & \quad \mathbf{z} = \frac{1}{F \cdot T} \sum_{i=1}^{F} \sum_{j=1}^{T} \mathbf{X}(i,j), \\[1ex]
\textbf{Excitation:} & \quad \mathbf{s} = \sigma\left( \mathbf{W}_2 \cdot \delta\left( \mathbf{W}_1 \cdot \mathbf{z} \right) \right), \\[1ex]
\textbf{Scale:}      & \quad \tilde{\mathbf{X}} = \mathbf{s} \odot \mathbf{X},
\end{align}
where $\sigma$ and $\delta$ are activation functions.

\subsubsection{Embedding and Self-Attention Blocks}
The embedding block (Fig.~\ref{fig:attention}) abstracts the channel–feature dimension (C $\times$ F) into tokens. The attention block then establishes long-range dependencies between these tokens. The attention mechanism is defined as \cite{DBLP:journals/corr/VaswaniSPUJGKP17}:
\begin{equation}
\text{Attention}(Q, K, V) = \mathrm{softmax}\left( \frac{Q K^\top}{\sqrt{d_k}} \right) V.
\end{equation}
A single attention block is used, but a multi-head strategy captures diverse channel relationships:
\begin{align}
\text{MultiHead}(Q, K, V) &= \text{Concat}(\text{head}_1, \dots, \text{head}_h) W^O, \\
\text{head}_i &= \text{Attention}(Q W_i^Q, K W_i^K, V W_i^V).
\end{align}

\subsubsection{Fully Connected Layer}
This module processes the attention output $\mathbf{Z} \in \mathbb{R}^{T \times D}$. Global average pooling is applied over the $T$ dimension, followed by layer normalization and two fully connected layers to produce the final output.

\subsection{Modality Fusion}
As shown in Fig.~\ref{fig:model}, EMG signals undergo the same multi-convolution and SE block processing as EEG. Fusion of EEG and EMG signals occurs within the embedding module.

\subsection{Copilot Framework}
\begin{figure}[t]
    \centering
    \includegraphics[width=0.95\linewidth]{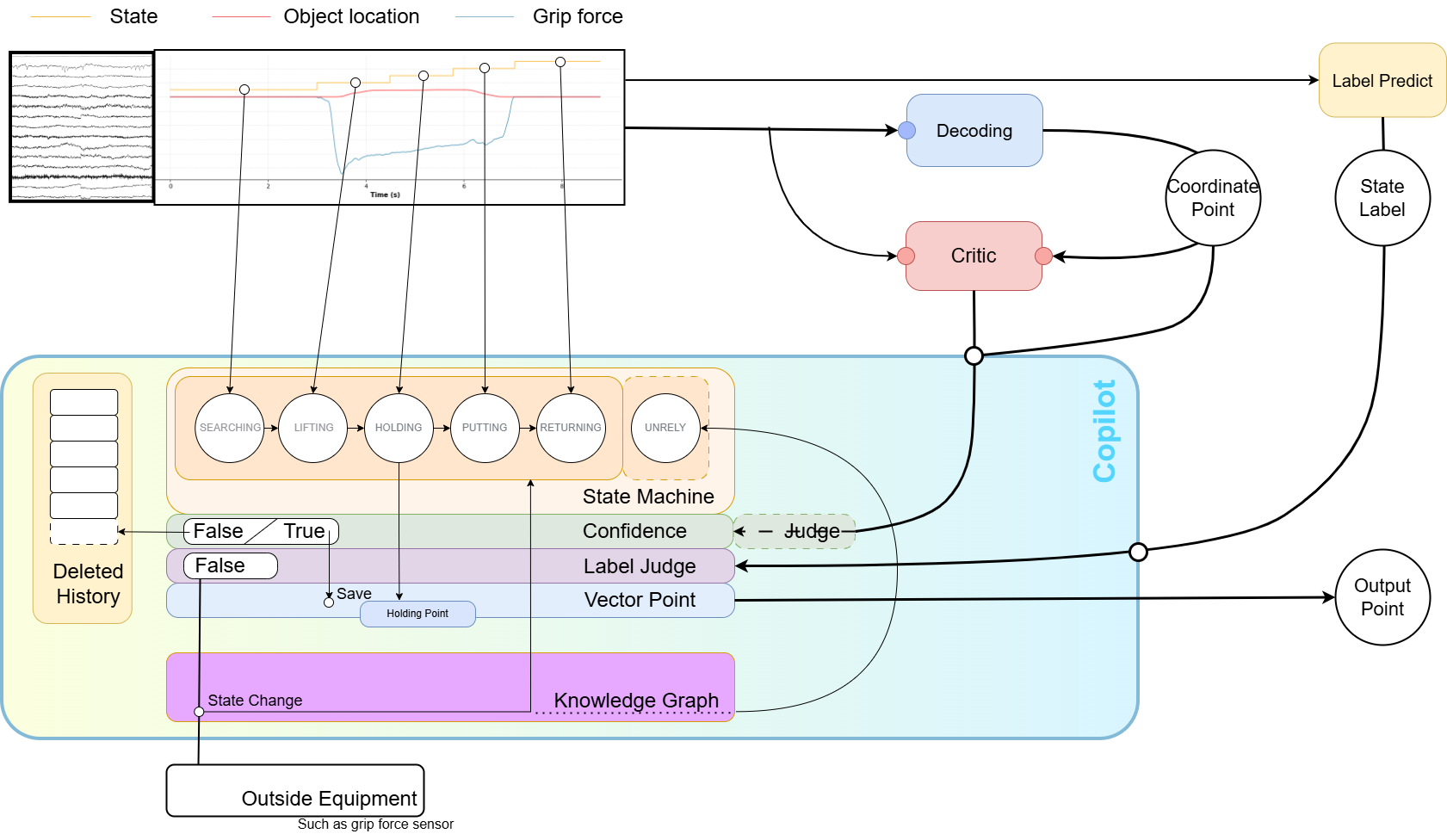}
    \caption{Structure of the copilot and decoding point filtering workflow. The grasp-and-lift movement is segmented into five states (SEARCHING, LIFTING, HOLDING, PUTTING, RETURNING), each with different confidence thresholds. UNRELY is a virtual state representing points misclassified as belonging to the current state, which are assigned a higher decoding threshold for filtering.}
    \label{fig:copilot_s}
\end{figure}

As depicted in Fig.~\ref{fig:copilot_s}, the copilot module integrates three components:
(1) a decoding model that predicts spatial coordinates,
(2) a label prediction model (identical structure) that classifies motion states, and
(3) a critic model that estimates confidence scores for each decoded point.
A knowledge graph interacts with external sensor information to drive state transitions within a finite state machine.

\section{Experiment Details}
\subsection{Experiment Design}
EEG and EMG signals from each grasp-and-lift trial served as decoding inputs. Considering MRCP latency and the transformer's sequence modeling capability, various window sizes (50–1000 samples) and delays (100–700 ms) were evaluated. Model input was a 3D tensor (batch $\times$ channel $\times$ time points), with outputs being min–max normalized 3D coordinates for the index finger and thumb (6 values total). The dataset was split following \cite{JainAnant2024EESI}: For each participant, we randomly selected 30 trials as the validation set and another 30 non-overlapping trials as the test set, the rest for training, using MSELoss.

For evaluating window sizes and delays, data from participant 4 was used. All delays and slicing were applied post-preprocessing to avoid artifact spread from filtering boundaries; segments with mismatched EEG and kinematics data after delay were discarded.

For within-subject model comparisons, the proposed model was tested against EEGNet, DeepConvNet, TCN, and a transformer baseline (EEG-only). The modified rEEGNet and rDeepConvNet from \cite{JainAnant2024EESI} were used. The multimodal version of our model used combined EEG–EMG input. Tests used a 250-sample window and 200 ms delay. Within-subject tests involved participants 3, 4, 5, 7, 9; cross-subject tests decoded participant 1's data using models trained on these five. Ablation studies validated model components.

Decoded trajectories were simulated on a 7-DOF Franka Panda robotic arm in MuJoCo. For each trial, the midpoint of decoded index finger/thumb trajectories was mapped to the arm workspace, converted to joint angles via inverse kinematics, and interpolated for continuous motion.

\begin{figure}[t]
 \centering
       \centering
        \includegraphics[width=\linewidth]{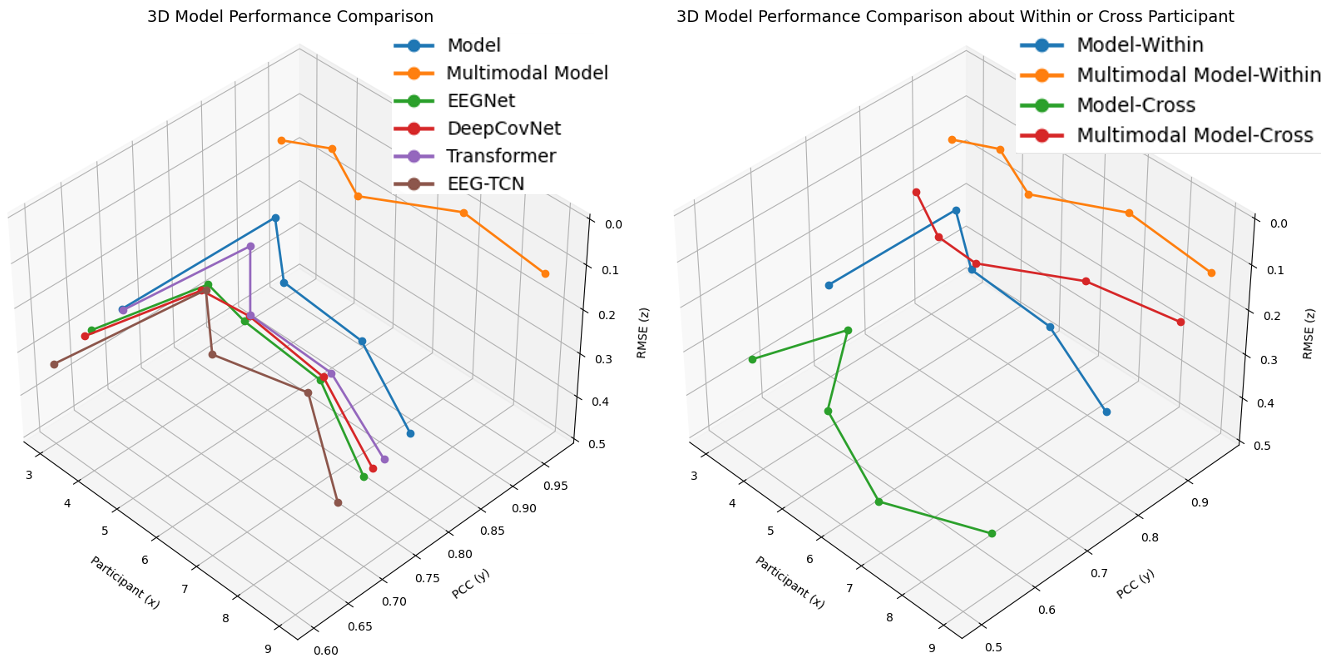}
        \caption{(Left) Within-subject decoding performance for different models (EEGNet, DeepConvNet, Transformer, EEG-TCN, EEG-only, and EMG–EEG fusion). (Right) Within-subject and cross-subject decoding performance for EEG-only and EMG–EEG fusion models.}
        \label{fig:overleaf2}
\end{figure}

\subsection{Copilot Implementation}
EEG preprocessing for the copilot was identical to previous experiments. The dataset was split differently: We reused the same trial-level train/validation/test split as previously for training the critic model, motion stage classifier, and state transition model.

\subsection{Performance Metrics}
\subsubsection{Pearson Correlation Coefficient (PCC)}
PCC measures linear correlation between decoded and actual 3D coordinate trajectories (thumb and index finger). The formula for two sequences $C^x$ and $C^y$ is:
$
PCC_{xy} = \frac{\sum_{i=1}^n (C_i^{x} - \bar{C^{x}})(C_i^{y} - \bar{C^{y}})}
{\sqrt{\sum_{i=1}^n (C_i^{x} - \bar{C^{x}})^2} \; \sqrt{\sum_{i=1}^n (C_i^{y} - \bar{C^{y}})^2}}.
$
The overall PCC is computed by concatenating all x, y, z outputs into a single vector.

\subsubsection{Root Mean Square Error (RMSE)}
RMSE quantifies the deviation between predicted and actual trajectories:
$
\text{RMSE}_{xy} = \sqrt{\frac{1}{N} \sum_{i=1}^{N} \left( C_i^{x} - C_i^{y} \right)^2 }.
$

\section{Results and Discussion}
\subsection{Model Comparison}
\subsubsection{Within-Subject Decoding}
As shown in Fig.~\ref{fig:overleaf2} (Left), except for the EEG–EMG multimodal approach, all EEG-only models showed notable inter-subject variability. 
The multimodal model remained stable across participants 3, 4, 5, 7, and 9, with PCC values of 0.94–0.97 and RMSE of 0.09–0.15. Participant 7 achieved the highest accuracy (PCC = 0.9707, RMSE = 0.0989), while participant 5 had the lowest (PCC = 0.9369, RMSE = 0.1447). In contrast, EEG-only models performed worse. Our proposed CNN–attention model performed best among them, achieving a PCC of 0.8728 and RMSE of 0.1908 for participant 4—only 0.08 PCC lower than the multimodal result. The transformer baseline followed (PCC = 0.8353, RMSE = 0.2208), while EEGNet, DeepConvNet, and TCN showed comparable results (PCC = 0.76–0.78, RMSE = 0.24–0.26). Participant 3 yielded the weakest EEG-only results, with PCC around 0.7 for both our model and the transformer, and RMSE near 0.3. TCN performed worst overall (PCC = 0.6036, RMSE = 0.3263).

These results indicate that our model provides a significant advantage over prior EEG-only models (EEGNet, DeepConvNet) for hand kinematics decoding, with an average PCC improvement of approximately 0.05, despite remaining sensitive to inter-subject variability.

\subsubsection{Cross-Subject Decoding}
As shown in Fig.~\ref{fig:overleaf2} (Right), cross-subject decoding resulted in lower PCC and higher RMSE than within-subject decoding.

For the EEG–EMG model, cross-subject decoding showed an average PCC decrease of ~0.09 and RMSE increase of ~0.08. The model trained on participant 9 generalized best (PCC = 0.9063, RMSE = 0.1781), showing the smallest drop from its within-subject performance. The model trained on participant 4 performed worst (PCC = 0.8388, RMSE = 0.2278).

The impact was more severe for the EEG-only model, with an average PCC drop of 0.23 and RMSE increase of 0.1. The best cross-subject performance came from the model trained on participant 4 (PCC = 0.6646, RMSE = 0.3139), still worse than the lowest within-subject result. The poorest performance was from the model trained on participant 7 (PCC = 0.4985, RMSE = 0.3894).

These findings suggest EEG–EMG multimodal signals are more robust to inter-subject variability. Both models showed a parallel trend between within-subject and cross-subject performance, implying that datasets yielding strong within-subject decoding also tend to support better cross-subject generalization under our deep learning architecture.

\begin{figure}[t]
 \centering
       \includegraphics[width=\linewidth]{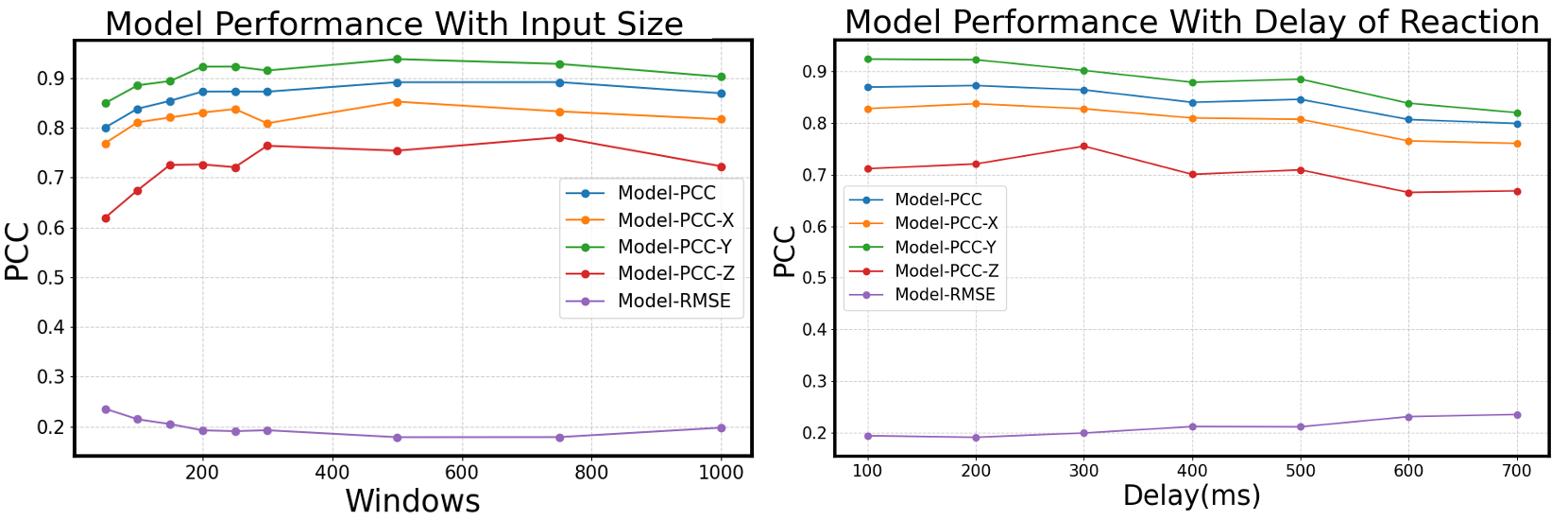}
       \caption{(Left) Model performance with varying input window sizes (50–1000 samples) for participant 4 with a 200 ms delay. Step size for the sliding window was one-fifth of the window length. (Right) Model performance with varying kinematic data delays (50–350 samples) for participant 4 with a fixed input length of 250 samples.}
       \label{fig:Figure_2}
\end{figure}

\subsubsection{Effect of Window Size and Delay}
We first study two practical factors that directly shape decoding quality: the EEG input window length and the temporal delay between EEG and kinematics. All results in this ablation are obtained in the EEG-only setting using participant 4, where PCC/RMSE are computed on the 3D midpoint trajectory between the thumb and index finger.

\textbf{Window size.} Fig~\ref{fig:Figure_2} (Left) shows a clear trade-off between temporal context and noise accumulation. As the window length increases, PCC improves steadily and reaches a turning point at 200 samples, indicating that short windows provide insufficient context for stable trajectory estimation. Beyond 200 samples, performance gradually declines without sharp oscillations, suggesting that overly long windows may dilute informative cues and introduce non-stationary interference. The lowest accuracy occurs at 50 samples (PCC = 0.8004), while the best performance is achieved at 750 samples (PCC = 0.8916). The axis-wise trends follow the same pattern, with the Y-axis consistently tracking the overall improvement most closely.

\textbf{Delay.} Fig~\ref{fig:Figure_2} (Right) evaluates the influence of kinematic delay. Performance peaks at a 200 ms delay and then degrades as the delay increases, consistent with the intuition that excessive lag weakens the time alignment between neural activity and motor execution. The best overall accuracy and error are obtained at 200 ms (PCC = 0.8728, RMSE = 0.1908), whereas the worst decoding is observed at 700 ms (PCC = 0.7993, RMSE = 0.2351). Across delays, axis-specific performance consistently follows Y $>$ X $>$ Z. The Y-axis achieves its highest PCC at 100 ms (0.9241), the X-axis peaks at 200 ms (0.8376), while the Z-axis is less stable and reaches its maximum at 300 ms (0.7554).

\begin{figure}
  \centering
   \includegraphics[width=1\linewidth]{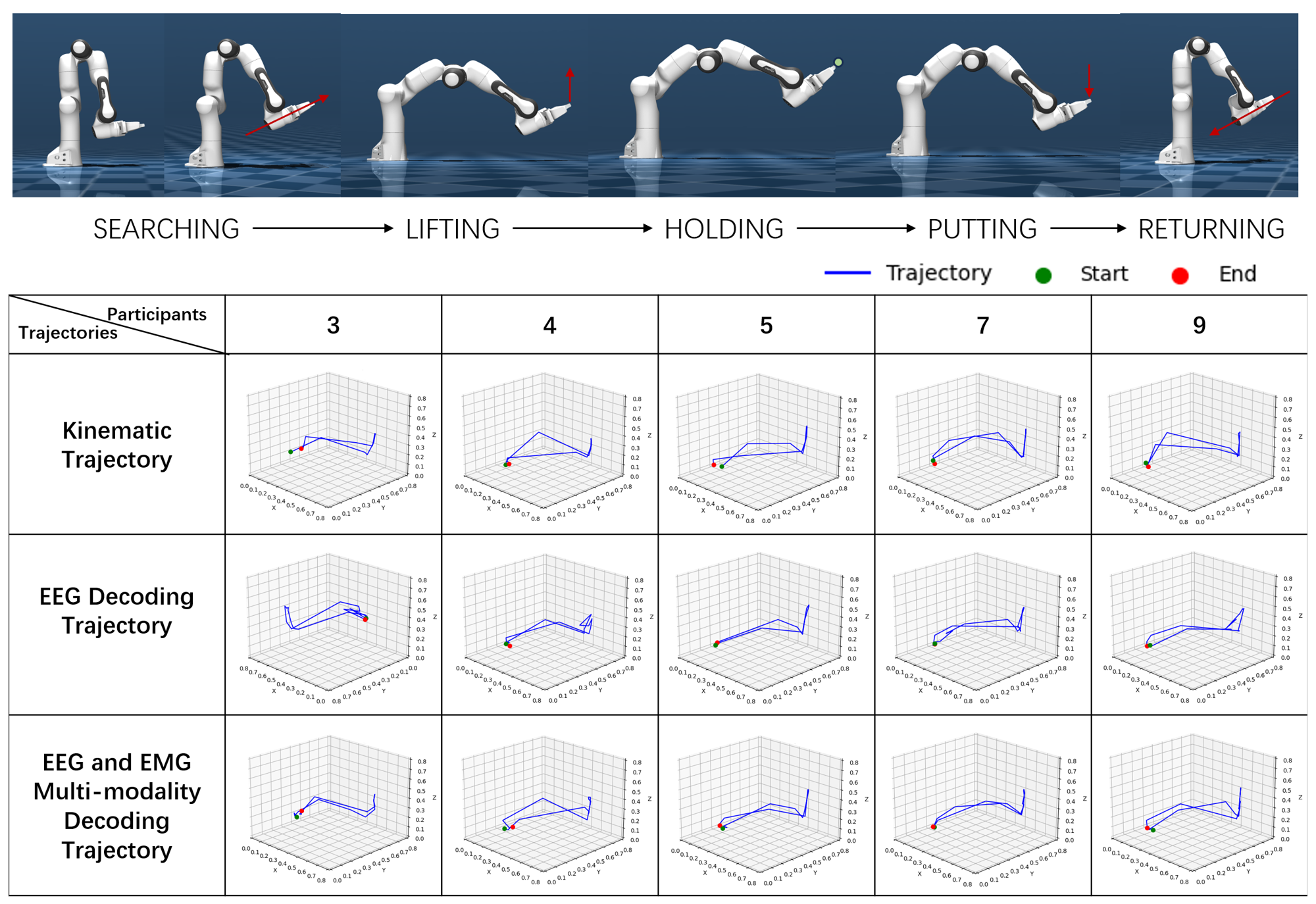}
        \caption{MuJoCo robotic arm simulation (top). Trials with relatively clear trajectories (high PCC) from participants 3, 4, 5, 7, and 9, reconstructing the grasp-and-lift movement. For participant 3, the trajectory was rotated 180° around the z-axis to resolve visual overlap (bottom).}
        \label{fig:Mujoco_T}
\end{figure}

\begin{figure}
    \centering
        \includegraphics[width=1\linewidth]{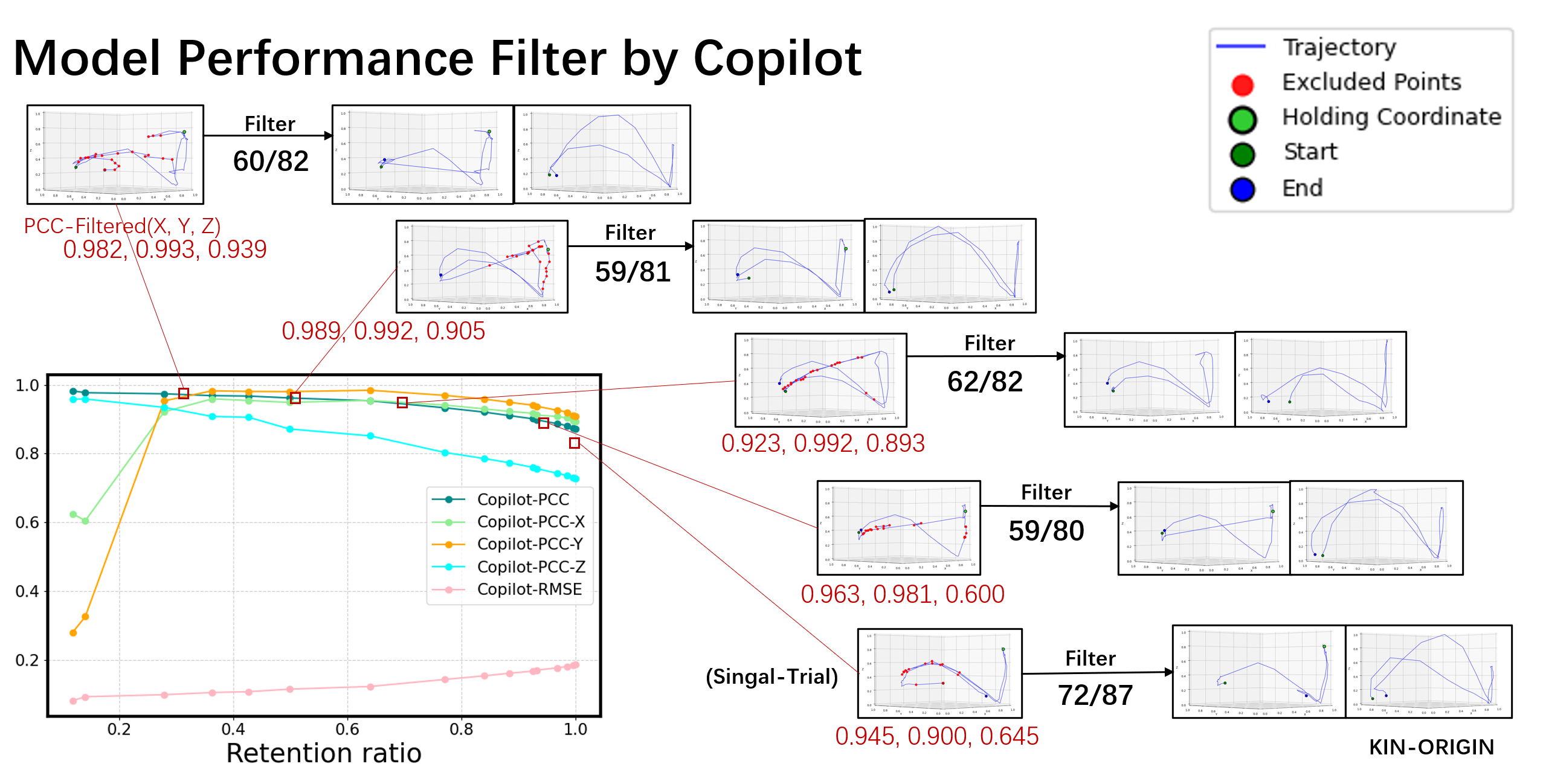}
        \caption{ Performance of the test dataset for participant 4 after copilot filtering, showing the proportion of retained points (bottom-left graph). Peripheral plots show trajectory changes before/after filtering for a randomly selected trial, compared to the ground truth. The red box indicates the overall PCC for a single filtered trial.}
        \label{fig:co}
\end{figure}


\subsection{Hand Kinematics Reconstruction on a Robotic Arm}
As shown in Fig.~\ref{fig:Mujoco_T}, EEG-only decoding occasionally produced trajectory reversals during all movement phases—particularly in grasping/return (participants 3, 7) and lift-hold stages (participants 3, 4, 9). Fluctuations during holding indicate residual precision limitations. Incorporating EMG significantly reduced the magnitude of reversals, especially for participants 3 and 4.

While the trajectories in Fig.~\ref{fig:Mujoco_T} appear relatively complete, this applies only to trials with very high PCC ($>$0.9). For most trajectories (PCC 0.83–0.88), reversals are more common and pronounced. To address this, we introduced the copilot filtering module, whose effect is shown in Fig.~\ref{fig:co}.

The graph shows that PCC continuously increased as the retention ratio decreased, up to a threshold. When the retained points dropped to 27.88\% (672/2410), performance metrics (including PCC) changed significantly and irregularly. This indicates that applying an appropriate confidence threshold in the copilot can enhance decoding quality, but excessive filtering is detrimental. The peripheral trajectory plots demonstrate that obvious reversals are filtered out, resulting in a clearer grasp-and-lift movement profile.

In summary, copilot-filtered, EEG-based decoding trajectories can roughly reconstruct the grasp-and-lift motion and generate a stable robotic arm trajectory. However, for low-accuracy trajectories, the copilot currently cannot correct points—it can only filter unreliable ones. Higher-precision trajectory reconstruction from non-invasive EEG still requires more robust decoding models and potentially improved hardware.

\section{Conclusion and Future Work}
This paper proposed a CNN–attention hybrid model for predicting hand kinematic trajectories from movement-related EEG or combined EEG–EMG signals. 
We also designed a copilot framework to filter decoding points, improving the quality of EEG-only trajectories. Future work will integrate additional sensors to provide assistance tailored to different motion patterns and explore more advanced methods for decoding EEG and controlling arms in a more robust fashion~\cite{yang_spiking_2025}. 

\bibliographystyle{IEEEtran}
\bibliography{main}
\end{document}